\documentclass[letterpaper, 10 pt, conference]{ieeeconf}

\usepackage{textcomp}
\usepackage{amsfonts,amssymb,amsmath}
\usepackage{booktabs}
\usepackage{stfloats}
\usepackage{url}
\usepackage{verbatim}
\usepackage{graphicx} 
\usepackage{cite}
\usepackage[version=4]{mhchem}
\usepackage{epsfig} 
\usepackage{mathptmx} 
\usepackage{times} 
\usepackage[T1]{fontenc}
\usepackage{hyperref}
\hypersetup{hidelinks,
    colorlinks=true,
    allcolors=black,
    pdfstartview=Fit,
    breaklinks=true}
\usepackage{multirow}
\usepackage{makecell}
\usepackage{diagbox} 
\usepackage{bm}
\usepackage{float}

\usepackage{cuted}
\usepackage{algorithmic}

\usepackage{tabularx}   
\usepackage{ragged2e}   

\usepackage[linesnumbered,ruled,vlined]{algorithm2e}
\usepackage{caption}
\usepackage{textcomp}
\usepackage{mathtools}
\usepackage{xcolor}
\newcommand{\projpage}{\href{https://avr-robot.github.io/}{\texttt{Project Page}}}

\newcolumntype{Y}{>{\centering\arraybackslash}X}

\IEEEoverridecommandlockouts
\title{\LARGE \bf AVR: Active Vision-Driven Precise Robot Manipulation with Viewpoint and Focal Length Optimization}

\author{
    Yushan Liu\textsuperscript{*1}, Shilong Mu\textsuperscript{*1}, Xintao Chao\textsuperscript{1}, Zizhen Li\textsuperscript{2}, Yao Mu\textsuperscript{3}, Tianxing Chen\textsuperscript{4}, Shoujie Li\textsuperscript{1}, \\ Chuqiao Lyu\textsuperscript{†1}, 
    Xiao-Ping Zhang\textsuperscript{1}, \textit{Fellow, IEEE}, Wenbo Ding\textsuperscript{†1}
}

\begin{document}

\maketitle

\begingroup
\renewcommand\thefootnote{}
\footnotetext{*These authors contributed equally to this work.}
\footnotetext{†Corresponding authors.}
\footnotetext{\textsuperscript{1} Tsinghua University}
\footnotetext{\textsuperscript{2} National University of Singapore}
\footnotetext{\textsuperscript{3} Shanghai Jiao Tong University}
\footnotetext{\textsuperscript{4} The University of Hong Kong}

\footnotetext{Project page: \url{https://AVR-robot.github.io}.}
\endgroup

\thispagestyle{empty}
\pagestyle{empty}

\begin{strip}
  \vspace{-18mm}
  \centering
  \includegraphics[width=1.0\textwidth]{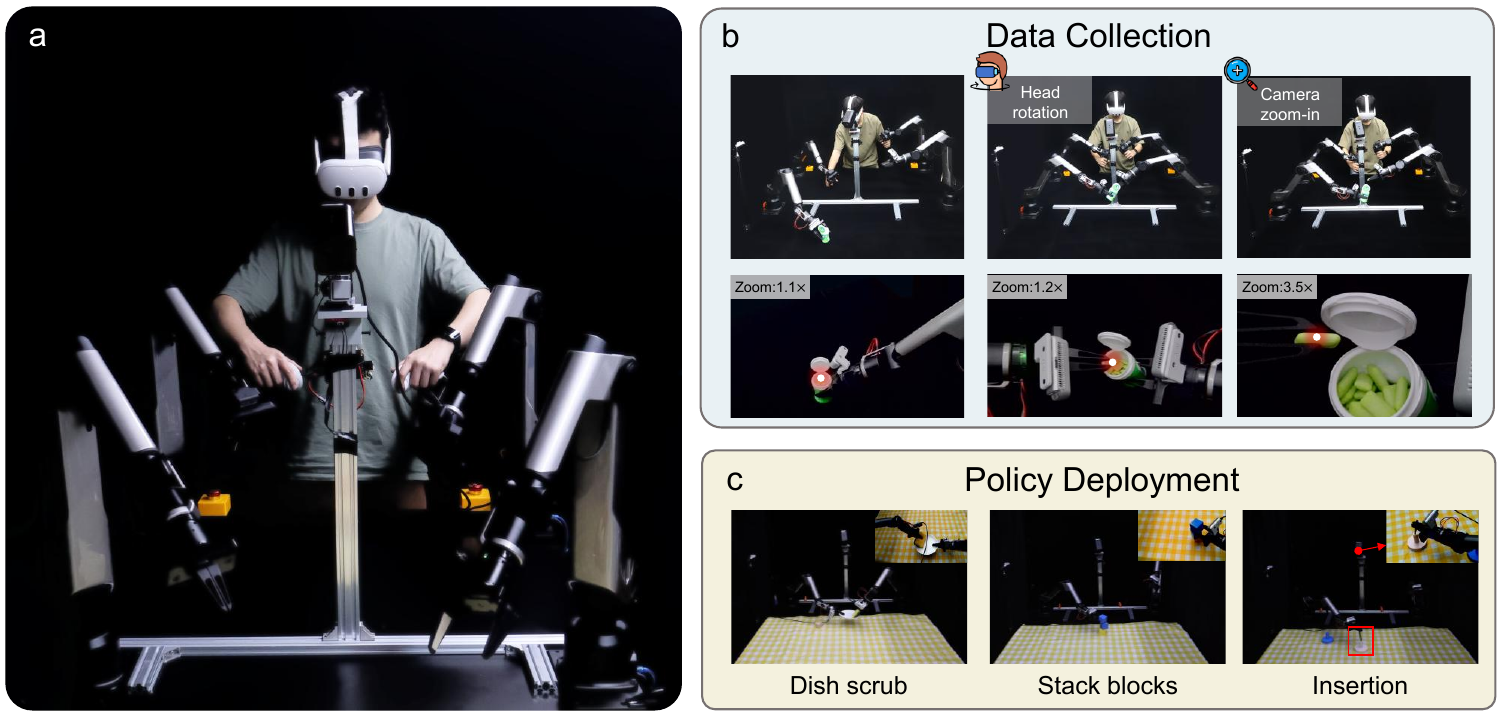}
  \captionof{figure}{\textbf{Overview of the AVR system}. (a) Hardware setup enabling bimanual control via VR controller or host–slave teleoperation. (b) Data collection with operator-controlled camera viewpoint (head pose) and focal length (zoom). (c) Example policy deployments: dish scrub, stack blocks, and insertion.}
  \label{fig:1}
\end{strip}
\begin{abstract}

Robotic manipulation in complex scenes demands precise perception of task-relevant details, yet fixed or suboptimal viewpoints often impair fine-grained perception and induce occlusions, constraining imitation-learned policies. We present AVR (Active Vision-driven Robotics), a bimanual teleoperation and learning framework that unifies head-tracked viewpoint control (HMD-to-2-DoF gimbal) with motorized optical zoom to keep targets centered at an appropriate scale during data collection and deployment. In simulation, an AVR plugin augments RoboTwin demonstrations by emulating active vision (ROI-conditioned viewpoint change, aspect-ratio-preserving crops with explicit zoom ratios, and super-resolution), yielding 5–17\% gains in task success across diverse manipulations. On our real-world platform, AVR improves success on most tasks, with over 25\% gains compared to the static-view baseline, and extended studies further demonstrate robustness under occlusion, clutter, and lighting disturbances, as well as generalization to unseen environments and objects.
These results pave the way for future robotic precision manipulation methods in the pursuit of human-level dexterity and precision.

\end{abstract}

\section{Introduction}

Imitation learning (IL) has emerged as a powerful paradigm for enabling dexterous robotic behavior in complex systems. Unlike reinforcement learning, IL eliminates the need for precise dynamics modeling or the manual design of reward functions by directly learning end-to-end control policies from expert demonstrations. This approach has achieved remarkable progress across a range of robotic manipulation tasks~\cite{osa2018algorithmic, ross2011reduction, ho2016generative, chi2023diffusion, mu2024robotwindualarmrobotbenchmark}. However, as task complexity increases, for example in cluttered environments or in operations that require high-precision control, conventional IL frameworks continue to face significant performance bottlenecks.

Data collection for imitation learning in dexterous manipulation is hampered by teleoperation noise and perceptual limitations, which together yield suboptimal demonstrations and throttle policy progress. Third-person teleoperation typically operates with limited situational awareness and delayed feedback, which often results in frequent errors or failed trials. This problem is especially evident in fine manipulation, and the inclusion of such demonstrations in training degrades the model’s ability to capture essential action patterns. At the sensing level, conventional camera configurations, which combine fixed external views with wrist-mounted cameras, rarely provide the high-fidelity cues needed for precision. Fixed cameras offer global context but lack sufficient spatial resolution in cluttered or precision-sensitive scenes. Wrist cameras, on the other hand, provide local views that are often occluded and subject to erratic viewpoints as the manipulator moves. Consequently, even with state-of-the-art IL algorithms and large-scale, high-quality demonstrations, success rates on complex tasks often plateau, indicating a bottleneck rooted not merely in data quantity but in the incomplete capture and representation of critical interaction details during data collection, which ultimately limits policy generalization and robustness.


Existing work has make progress in VR-based teleoperation \cite{cheng2024opentelevisionteleoperationimmersiveactive, chuang2024activevisionneedexploring, xiong2025visionactionlearningactive} and large-scale dataset building \cite{bu2025agibot, mittal2023orbit, wang2023robogen, khazatsky2024droid}. However, acquiring high-quality real-world demonstrations remains costly and time-consuming, and, as noted above, simply scaling data often yields diminishing returns, with success rates saturating on fine-grained manipulation tasks.


In contrast, human operators naturally leverage attention mechanisms to dynamically filter and focus perception on task-relevant regions and features. This adaptive allocation of sensory resources enables humans to consistently perceive and react to fine details, even under high cognitive load and visual clutter. Inspired by this, we posit that equipping robots with a similar capability to \textbf{actively focus on and magnify critical interaction details} during demonstration collection can effectively mitigate the aforementioned limitations.

To this end, we propose \textbf{Active Vision-Driven Robotics (AVR)}, a bimanual manipulation system that integrates an active vision module capable of dynamically adjusting both camera viewpoint and optical zoom. The framework consists of the following components:

\begin{itemize}
    \item \textbf{Autonomous Optical Zoom Camera}: A controllable zoom camera to enable real-time magnification of task-relevant regions during demonstration. The recorded zoom information and video are jointly used as inputs for policy learning, providing an additional modality focused on fine-detail perception.
    \item \textbf{Egocentric VR with Head-Tracked View Control}: Stream the optical zoom camera to an VR head motion device(HMD) and map the operator’s head pose to a 2-DoF gimbal for real-time viewpoint adjustment, keeping the target centered and yielding higher-fidelity, task-relevant demonstrations.
\end{itemize}

We evaluate our AVR framework across both simulation and real-world bimanual platform. Results demonstrate that our approach consistently improves task success rates across diverse scenarios, especially under cluttered or precision-demanding conditions. Notably, AVR achieves up to 30\% improvement in success rate compared to state-of-the-art baselines under limited data conditions, validating its effectiveness in enhancing demonstration quality and policy performance through active visual attention.

\section{Related Work}

\subsection{Active Vision for Robot Policy Learning}
Active vision have been widely applied in robotics, especially for robot policy learning\cite{zeng2020view, wang2025observeactasynchronousactive}. 
Recent work mostly focus on "attention guidance" -- either controlling camera viewpoints to reduce occlusion and field-of-view limitations, or jointly learning a sensory (viewpoint) policy with the motor policy in partially observable settings -- thereby steering perception toward task-relevant regions and mitigating dilution in global views. \cite{chuang2024activevisionneedexploring, rakita2019remote}
In parallel, encoder-centric approaches (e.g., voxelized or two-stream encoders) emphasize spatially selective processing to highlight manipulation-relevant areas without explicitly moving the camera\cite{shridhar2022peract, shang2023active}.
However, these strategies often under-represent fine, manipulation-critical cues, a gap made evident when fixed or suboptimal viewpoints induce occlusions and miss small details. 
In this work, we extend additional detail observation that dynamically adjusts viewpoint and zoom to capture critical details, enabling more precise and robust manipulation. 

\subsection{Learning from Details}

Learning from details is a recurring strategy for boosting perception and decision quality across domains. In fine-grained recognition, models mine subtle cues via region-level attention or localized feature interactions\cite{fu2017look, yang2018learningnavigatefinegrainedclassification}; In medical imaging, zoom-in pipelines explicitly crop high resolution patches around suspicious findings to capture lesion-level evidence\cite{wang2017zoominnetdeepmininglesions}; For small-object detection, surveys and task-driven super-resolution demonstrate that recovering fine structure materially improves recognition in low resolution or cluttered scenes\cite{feng2023deep}.

Robotics has also benefited from detail-centric perception\cite{chuang2025lookfocusactefficient}. Building on these insights, we go beyond implicit attention by introducing an explicit detail-observation modality. In practice, the system dynamically adjusts viewpoint and zoom to deliver high-resolution local evidence to the policy, thereby improving precision and robustness in fine manipulation.

\begin{figure}[thpb]
  \centering
  \includegraphics[width=1.0\linewidth]{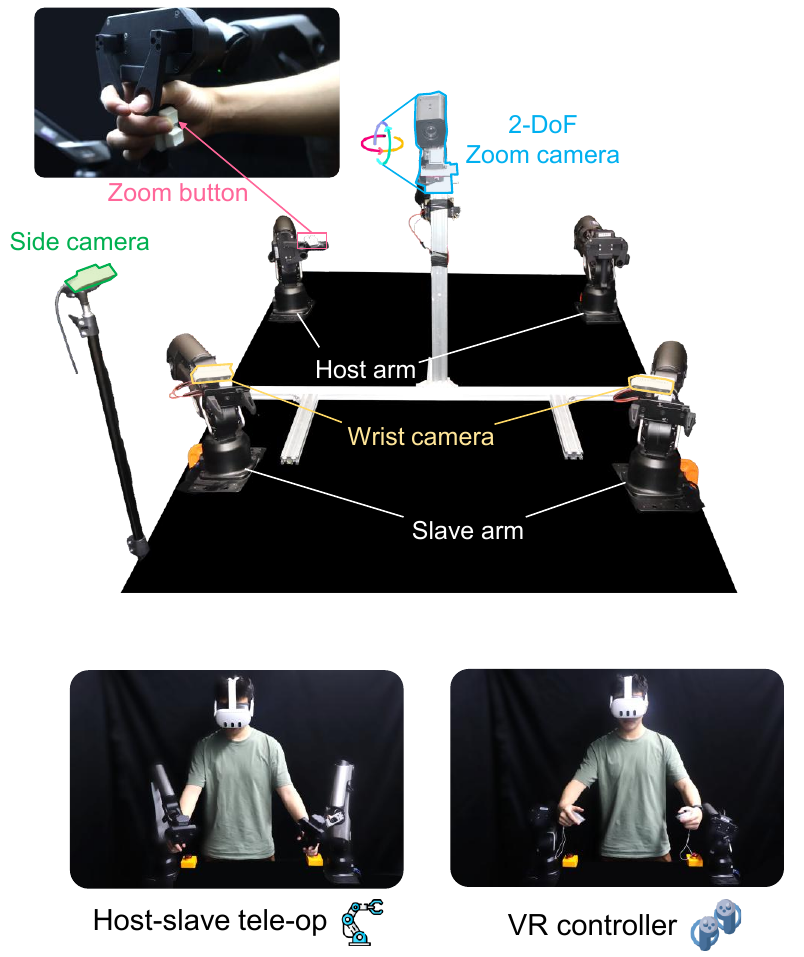}
  \caption{AVR hardware platform. A 2-DoF zoom camera provides full-workspace coverage and adjustable focal length, complemented by a side camera and wrist-mounted cameras. System supports two control modalities: host–slave teleoperation and VR controller, each enabling real-time zoom control of the active camera (host–slave: dedicated zoom button; VR: controller button mapping).}
  \label{fig:2}
\end{figure}

\section{AVR System Architecture}

\subsection{Hardware System}

Fig.~\ref{fig:2} shows the hardware architecture of our AVR system. The manipulation platform consists of two 6-DoF Galaxea robotic arms, each equipped with a parallel-jaw gripper\cite{GalaxeaA1}. Three Intel D435i depth cameras are positioned to capture side and wrist view, ensuring complete observation of the manipulation workspace. An active vision module is mounted on top of the platform, comprising a motorized-zoom industrial camera and a 2-DoF gimbal. It captures real-time human head motion with live visual feedback during teleoperation, which facilitates fine-grained perception and the recording task details.


\subsection{Teleoperation for Data Collection}

To collect high-quality real-world manipulation data, we build a teleoperation framework with two control modes: (i) host–slave joint-angle mapping and (ii) 6-DoF end-effector control via VR controllers. This dual setup accommodates diverse tasks and operator preferences.
For perception, we provide a VR-projected egocentric view with head-tracked active viewpoint control, closing a low-latency observation–action loop.
We also add active zoom: a button–zoom mapping (keyboard button or VR controller) enables real-time focal adjustments during bimanual operation; at deployment, the policy autonomously modulates zoom to retain detail-rich observations.

\textit{\textbf{Pose-to-Gimbal Mapping}}: Prior studies have demonstrated that 2-DoF gimbals can effectively approximate human head motion while maintaining adequate coverage of the operational workspace
 \cite{nakanishi2020towards}. We leverage head motion estimates from HMD to capture head orientation, employing a One-Euro filter\cite{casiez20121} for adaptive noise mitigation, and ultimately outputting gimbal rotation angles:
\begin{equation}
\bm{\theta}_\textit{t} = \Pi\!\left(\;  \bf{R}\, (\,\mathcal{F}_{\text{1-Euro}}( \tilde{\bm q}_{\textit{t}}\,) ) \right)
\end{equation}
where \(\tilde{\bm q_\textit{t}}\,\in\,\mathbb{S}^3\) is the raw unit quaternion at time $t$, \( \mathcal{F}_\text{1-Euro}\) is the One-Euro filter defined on \(\mathrm{SO}(3)\) with dynamically adjusted parameters:
\begin{equation}
f_t=f_{\min}+\beta\left\|\dot{\omega}_{\,t}\right\|,\quad
\alpha_{\,t}=\frac{2\pi f_t\,\Delta\,t}{1+2\pi f_t\,\Delta\,t},
\end{equation}
where \(\dot{\omega}_{\,t}\) is a low-pass estimate of angular velocity and \(\Delta t\) is the sampling interval. \(\mathbf R(\cdot)\) maps quaternion to rotation matrix. $\Pi(\cdot)$ extracts yaw/pitch from rotation matrix: letting $\bm k=\bf R^{\,\mathrm{gimbal}}_{\,\textit{t}}\,\mathbf e_\textit{z}$, with \(\bm k = [k_x,k_y,k_z]^\top\), \(\mathbf e_\textit{z}=[0,0,1]^\top\):
\begin{equation}
\begin{bmatrix}\theta^{\text{yaw}}_{\,t}\\ \theta^{\text{pitch}}_{\,t}\end{bmatrix} =\Pi(\bf R^{\,\mathrm{gimbal}}_{\,\textit{t}})=
    \begin{bmatrix}
    \arctan(k_x,\,k_z)\\[2pt]
    \arctan\!\big(-k_y,\,\sqrt{k_x^2+k_z^2}\big)
    \end{bmatrix}
\end{equation}

Compared with direct pose-to-gimbal mapping, the filter-based pipeline suppresses jitter more effectively at rest and delivers smoother tracking during rapid head motions, resulting a more stable view for teleoperation.

\textit{\textbf{Camera-to-HMD Projection}}: To ensure a clear operational viewpoint, we adopt spherical rendering with an average end-to-end latency of approximately 80 ms. The live 60 fps camera stream is projected onto a \(110^\circ\) curved surface at a 1 m radius around the HMD viewpoint. By preserving vestibular–visual consistency, this spherical rendering can effectively mitigate motion sickness\cite{schwarz2021lowlatencyimmersive6dtelevisualization}. Combined with mentioned gimbal control, it provides a high-fidelity immersive perspective for precise teleoperation tasks.

\textit{\textbf{Button-to-Zoom Camera Control}}. To better focus on and record fine interaction details during data collection, we equip both teleoperation modes with a button-to-zoom mapping for real-time control of the active vision camera. In the host–slave teleoperation, the operator uses a keypad on the teach pendant to zoom in/out; in the VR mode, controller buttons are mapped to camera zoom, enabling simultaneous bimanual operation and real-time adjustment of field of view and focal length. At deployment, the camera’s focal length is autonomously adjusted according to the policy’s learned zoom behavior, ensuring detail observation during from collection to execution.

\begin{figure*}
    \centering
    \includegraphics[width=1\linewidth]{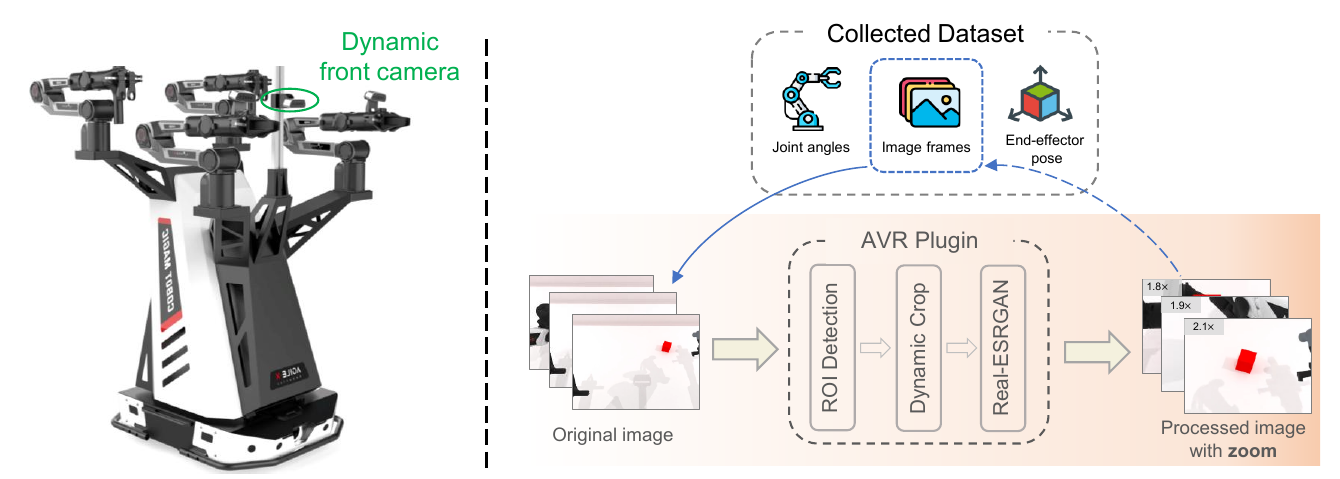}
    \caption{AVR Plugin in simulation. We take the front camera view from RoboTwin collected dataset as input, get detailed observation by ROI detection, aspect-ratio crop with zoom, and Real-ESRGAN super reconstruction. The processed images with zoom are appended to dataset as additional detailed observation for policy. }
    \label{fig:plugin_framework}
\end{figure*}

\subsection{Learning Policy}

We designed a policy network based on Diffusion Policy\cite{chi2023diffusion}. By leveraging external and active vision observations, along with proprioceptive state, the network predicts control actions for the system.

At each time step \(t\), the policy receives the current RGB image observations \(\mathcal{ I}_t = \{\bm I^i_t\}_{i=1}^4 \in \mathbb{R}^{H\times W\times C}\) as visual input, comprising three external viewpoints (wrist camera 1, 2 and side camera 3) and an active viewpoint (camera 4). We use pretrained DINOv2\cite{oquab2024dinov2learningrobustvisual} ViT as visual encoder for each \(\mathcal{\bm I}_t^i\), which produces \(16\times22\) tokens as scene representation. The proprioceptive state \(\bm p \in \mathbb R^{19}\), includes the end-effector poses (position and quaternion) of two arm (\(\in \mathbb R^{7\times 2}\)) with two gripper (\( \in \mathbb R^2\)), 2-DoF gimbal angles (\( \in \mathbb R^2\)), and camera zoom (1 scalar). The policy outputs a sequence of future actions \(\bm a_t = \{a_{t+1}, \dots, a_{t+n}\} \in \mathbb{R}^{n \times 19}\), where each \(a_{t+k}\) comprises two arm end-effector poses with gripper widths, gimbal angles, and zoom setting (all expressed in the world frame).

\begin{figure}
    \centering
    \includegraphics[width=1\linewidth]{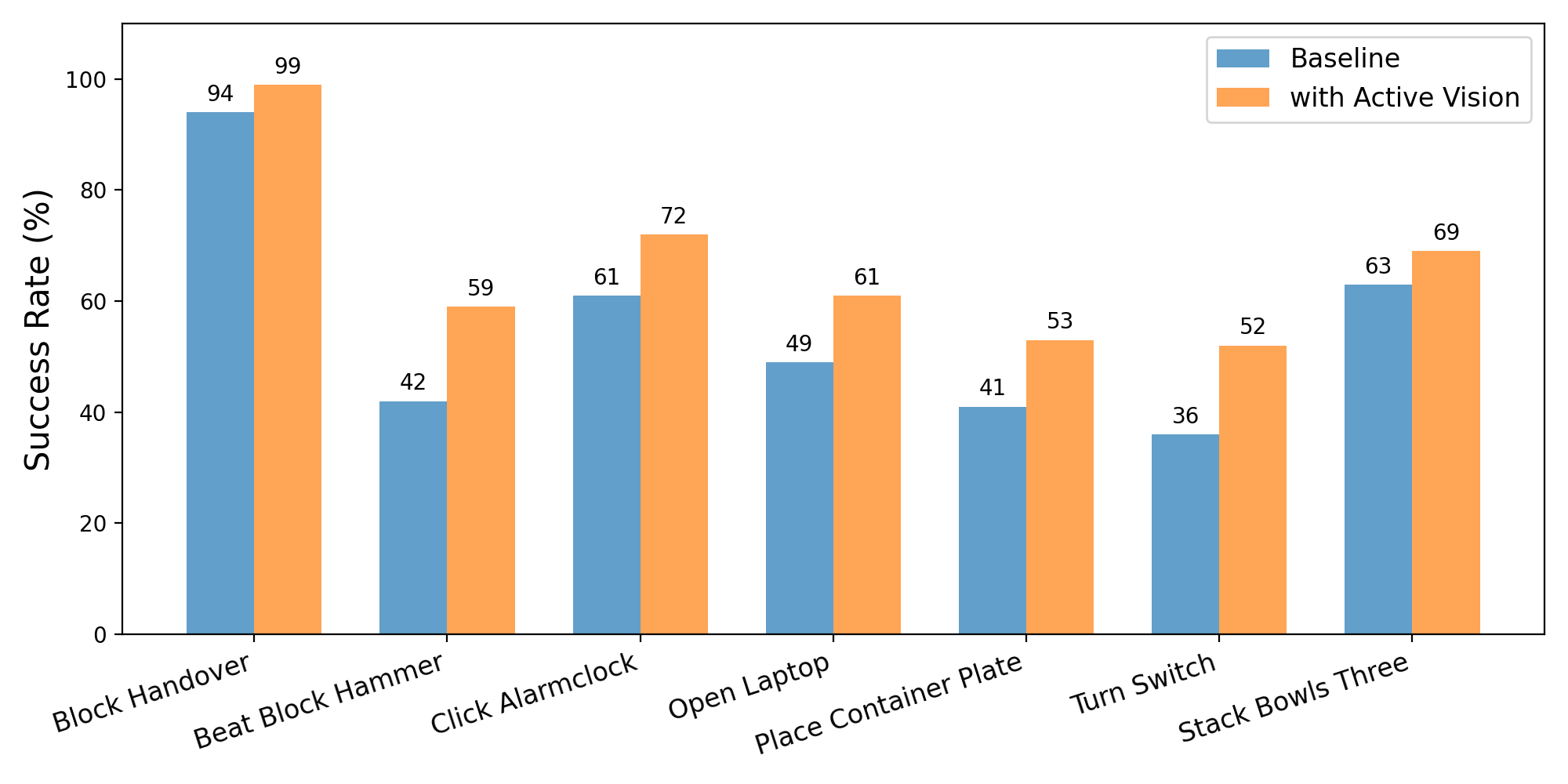}
    \caption{Comparison of tasks success rates between baseline and Active Vision in simulation, which indicate that detail observation help policies better execute manipulation.}
    \label{fig:result-sim}
\end{figure}

\section{Experiment}

To comprehensively evaluate the capabilities of the AVR framework, we designed a series of experiments spanning both simulated and real-world environments. These experiments aim to assess the system's performance across various robotic manipulation tasks, ranging from basic pick-and-place operations to high precision tasks requiring complex control strategies.

\begin{table}[htbp]
    \centering
    \caption{Trials needed to reach 50 successful demonstrations per task (successes/total).}
    \label{tab:teleop-times}
    \begin{tabularx}{\columnwidth}{@{}lYYY@{}}
        \toprule
        \textbf{Task} & ALOHA  & VR-control & AVR  \\
        \midrule
        Pick-place         & 50/50 & 50/50 & 50/50 \\
        Dish scrubbing     & 50/52 & 50/53 & 50/50 \\
        Fold cloth         & 50/51 & 50/53 & 50/50 \\
        Place cup on rack  & 50/50 & 50/52 & 50/50 \\
        \midrule
        Block stacking     & 50/56 & 50/60 & 50/52 \\
        Grasp chewing gum  & 50/61 & 50/57 & 50/55 \\
        Insert screwdriver & 50/69 & 50/58 & \textbf{50/53} \\
        \bottomrule
    \end{tabularx}
\end{table}

\begin{figure}
    \centering
    \includegraphics[width=1.0\linewidth]{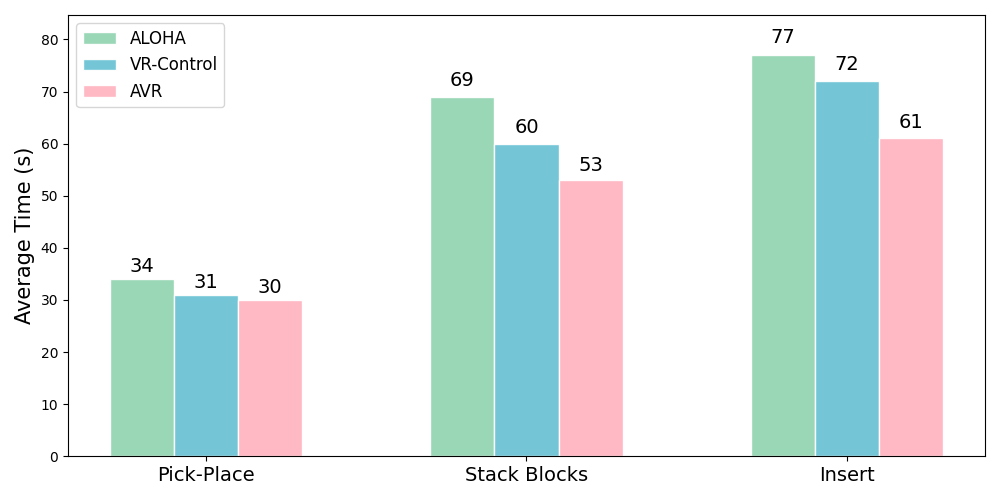}
    \caption{Average task completion time under different teleoperation settings. Across representative tasks, operators using AVR complete demonstrations in less time compared to ALOHA and VR-control, indicating fewer failed attempts and more efficient one-shot executions.}
    \label{fig:teleop}
\end{figure}

\subsection{RoboTwin-based Simulation Evaluation}
To evaluate whether detailed observation benefits policy performance, we extend RoboTwin~\cite{chen2025robotwin20scalabledata}, a simulation data collection platform, with an active vision module (named AVR Plugin) and conduct diverse manipulation tasks. Shown in Fig. \ref{fig:plugin_framework}, we perform offline processing on collected dataset containing RGB images, joint angles, and end-effector poses. By extracting the front camera view, which covered the entire manipulation workspace, as AVR Plugin's input. First, we applied task-conditioned ROI detection on target objects to simulate \textbf{dynamic camera viewpoints}; Then we executed an aspect-ratio-preserving crop and computed the relative zoom ratio to simulate \textbf{zoom variation}. To match the expected detail level after zooming, we applied Real-ESRGAN\cite{wang2021realesrgantrainingrealworldblind}, a state-of-the-art super resolution algorithm, to super-resolve the cropped region back to the original image resolution. The processed images, together with their corresponding zoom ratio, were integrated as additional details observation. 

We deployed 50 expert demonstration for each task, trained the model based on Diffusion Policy, and evaluate their performance in simulation. As shown in Fig. \ref{fig:result-sim}, our approach yielded 5\%-17\% increase across baseline, indicating that detail-aware observations help policies better understand and execute manipulation goals.

\subsection{Real Robot Performance}

\begin{figure*}[thpb]
  \centering
  \includegraphics[width=1.0\linewidth]{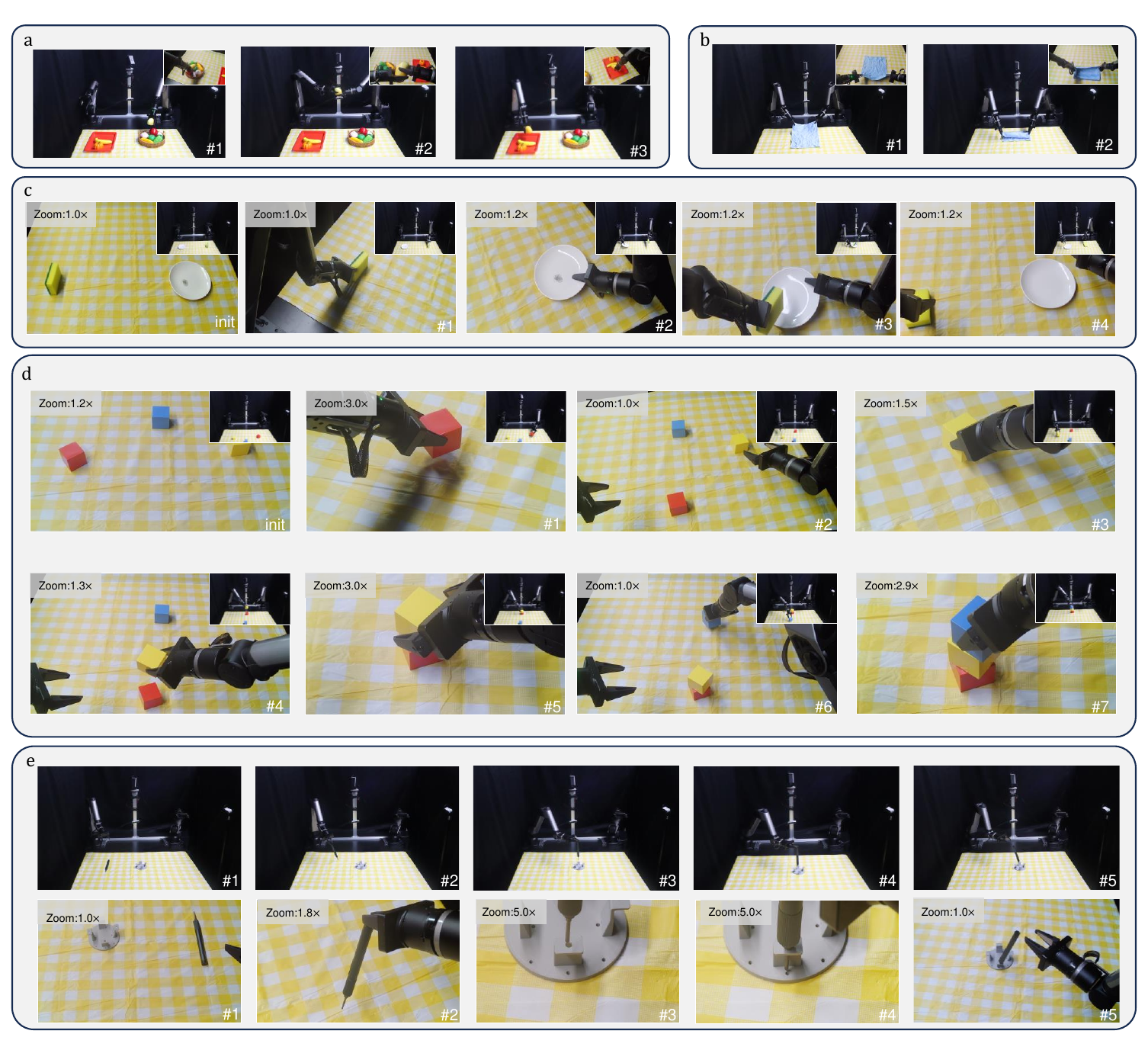}
  \caption{Deployment of various manipulation tasks. (a) Pick-and-place of objects with varied shapes. (b) Folding fabric with coordinated bimanual manipulation. (c) Dish scrubbing with a controlled wiping motion. (d) Block stacking requiring precise alignment. (e) Inserting a screwdriver tip into a hole for assembly. We provide first-person views from the active vision camera at various stages of each task, capturing changes in viewpoint and real-time focal adjustments.
}\label{fig:4}
\end{figure*}

We designed a suite of real-world bimanual manipulation tasks on our hardware platform to assess the effect of active vision on overall performance, espcially precision manipulation. Representative real-robot deployments are illustrated in Fig. \ref{fig:4}: (a) grasp a mango, perform a handover, and place it on a plate; (b) fold a towel once; (c) scrub a stained round dish; (d) sequentially grasp and stack three cubes (edge length 5 cm); and (e) grasp a small screwdriver (shank diameter 1 cm, head 0.5 cm) and accurately insert it into a fixed hole of 0.75 cm diameter. We further compare data collection efficiency and reliability across different teleoperation settings. 
Fig. \ref{fig:teleop} reports the average completion time across representative tasks, where operators using AVR complete trajectories more efficiently with fewer erroneous attempts, resulting in more consistent one-shot executions. 
Table \ref{tab:teleop-times} summarizes the number of failed trials during data collection, showing that AVR significantly reduces failures compared to conventional settings. 
This not only indicates improved data quality, but also demonstrates that even novice operators can more easily collect reliable demonstrations with our framework.
For each task, we collected 50 demonstrations under our hardware platform and trained policies using a diffusion policy framework to evaluate performance across conditions. 
Augmenting both data collection and deployment with AVR yields higher success rates on most tasks, substantially improving manipulation performance.

We further analyzed the differential impact of viewpoint and focal-length adjustments on task success. Prior works~\cite{morrison2019multiviewpickingnextbestviewreaching, lin2024perception} indicate that optimizing camera viewpoint improves grasping in clutter, whereas increasing magnification/ zoom benefits fine motor tasks (sometimes at the cost of completion time). Our results align with these trends: for typical pick-place tasks (e.g., handover, towel folding, dish scrubbing), maintaining target visibility demands substantial top-camera viewpoint changes, while only minor zoom (\(\le 2\times\)) is required; consequently, focal-length variation offers limited marginal gains. In contrast, for precision tasks (e.g., three-block stacking and small-hole insertion), viewpoint changes alone are insufficient to localize boundaries or apertures accurately. Adding dynamic focal length to obtain high-resolution local detail markedly improves alignment and positioning at critical phases, leading to higher success rates.

In summary, coarse pick-and-place tasks benefit primarily from viewpoint control, while precision manipulation benefits primarily from zoom. When the policy is equipped to actively acquire high-resolution detail at key moments, together with adequate viewpoint coverage, reliability and robustness are significantly improved on complex, high-precision tasks.


    
    

\begin{figure*}
    \centering
    \includegraphics[width=1\linewidth]{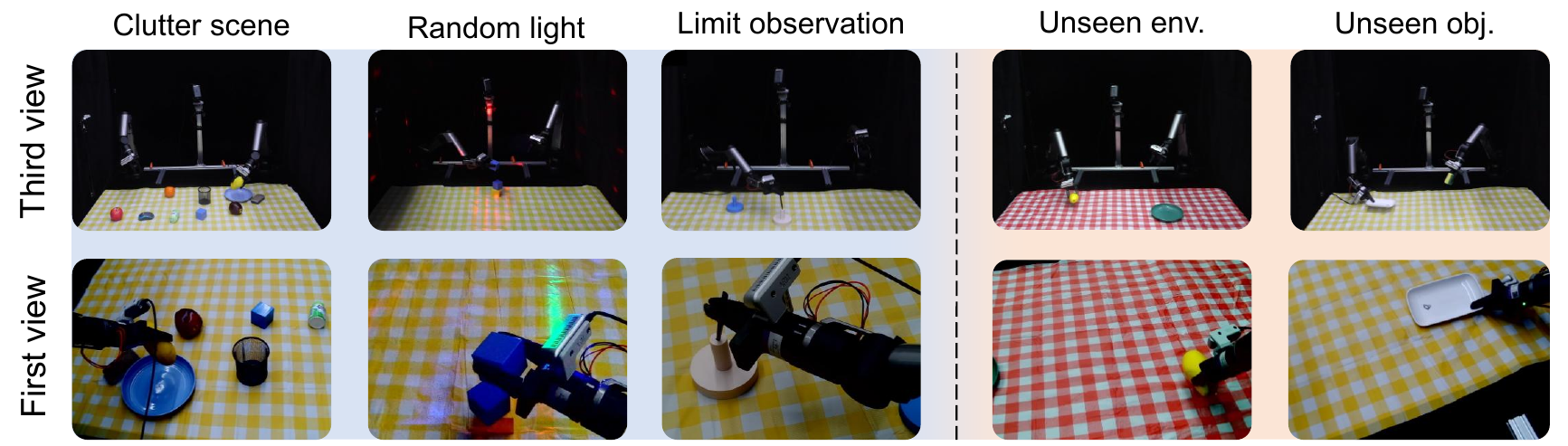}
    \caption{Extended experiments. We assess policy robustness and generalization on the base tasks under four conditions: cluttered scenes and random lighting (left), and transfer to unseen environments and unseen objects (right), with third and first real-time view.}
    \label{fig:extend}
\end{figure*}

\begin{figure}
    \centering
    \includegraphics[width=1\linewidth]{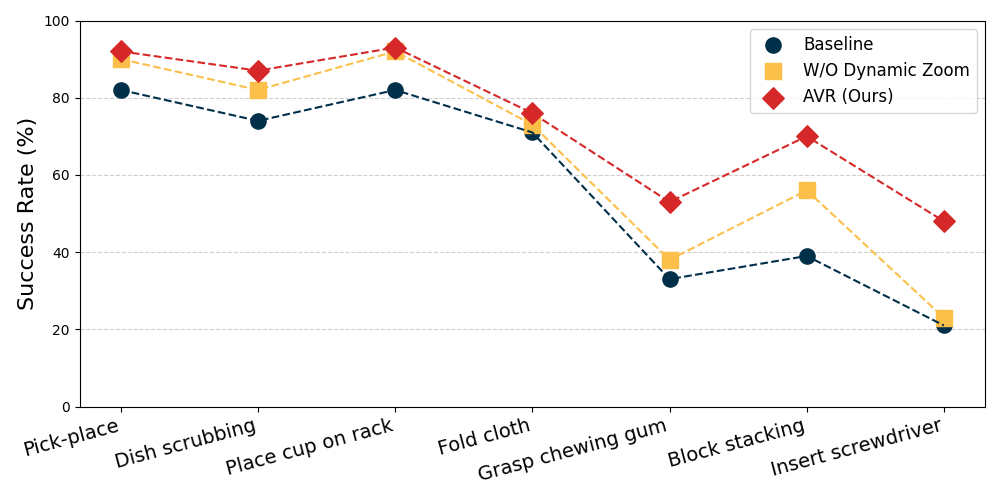}
    \caption{Ablation study for active vision. We compare static view, dynamic viewpoint only, and our AVR (dynamic viewpoint with focal length) on several tasks. AVR consistently achieves higher success rates, especially on fine-grained tasks.}
    \label{fig:real}
\end{figure}

\begin{figure}
    \centering
    \includegraphics[width=1\linewidth]{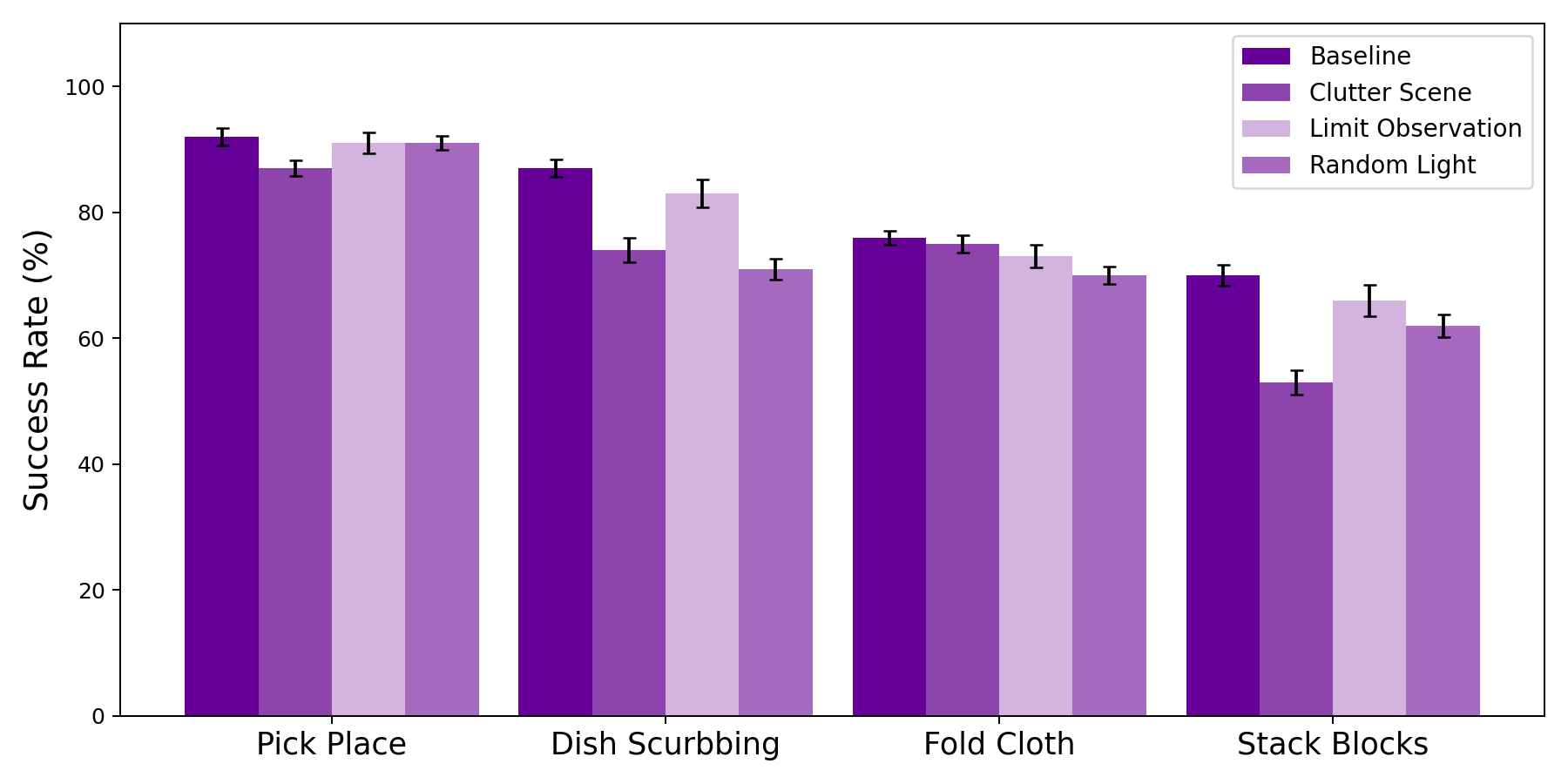}
    \caption{Robustness under perturbations. We choose four typical manipulation tasks with three disturbance: cluttered scene, random light and limit observation. Result shows that occlusion induces the largest degradation on visually demanding tasks, highlighting the importance of detail-aware observation for robustness.}
    \label{fig:robustness}
\end{figure}

\subsection{Extended Analysis}

In this section, we evaluate our method through multiple quantitative experiments, including ablation study, robustness assessment, and generalization validation. Some experimental results are presented in Fig. \ref{fig:extend}, and more deploy videos can be found on our \projpage.

\textbf{Ablation Study}: To quantify the contribution of our AVR framework, we conduct ablations on visual observation, comparing three settings: (i) no active vision, (ii) dynamic viewpoint only (traditional active vision method), and (iii) dynamic viewpoint and focal length (our method). Fig. \ref{fig:real} shows that with additional detailed observation, success rates rise across all tasks, notably on fine-grained manipulations. 


\textbf{Robustness}: Robust control under realistic perturbations is a core criterion of policy competence. We evaluate four manipulation tasks under three disturbance families: (i) limited observability via viewpoint occlusion, (ii) cluttered scene, and (iii) illumination disturbance, while randomizing object start poses. Fig. \ref{fig:robustness} summarizes the success rates (mean ± 95\% CI). Tasks that rely more on visual observation (e.g., cloth folding) show more degradation under occlusion, highlighting the importance of detailed observation in maintaining robustness.

\textbf{Generalization}: A policy’s ability to operate in unseen environments and with unseen objects is key to distinguishing task understanding from mere trajectory cloning. We evaluate generalization on multiple tasks under two settings: (i) unseen environments, where we vary scene layout and tabletop background; and (ii) unseen objects, where targets are replaced by new instances of the same category. Our method maintains higher success in unseen settings and shows smaller from seen to unseen, demonstrating robust cross-environment and cross-instance generalization enabled by viewpoint-and-focal length active vision.

\section{Conclusion and Future Work}

In this work, we introduce the AVR framework, which leverages dynamic viewpoint and focal length adjustments in active vision to enhance precise manipulation. The system provides an intuitive teleoperation experience and a reliable data collection workflow, ensuring consistent dynamic viewpoint and zoom adjustments that contribute to stable operation and improved control during precision tasks. It demonstrates improved precision in kinds of manipulation tasks. Simulation and real-world experiments show that AVR improves task success rates by 5\% - 17\%, with more than \textbf{25\%} increases in precision for precision tasks, significantly outperforming conventional imitation learning methods. Extended experiments confirm that our AVR framework, by enabling detailed observation and learning, yields substantial policy gains even when only limited data are available. The resultant policies demonstrate enhanced robustness under occlusion, clutter, and lighting disturbances, exhibiting smaller performance declines. Furthermore, they maintain higher success rates and smaller performance gaps when deployed in unseen environments and on novel objects, showcasing superior cross-environment and cross-instance generalization and underscoring the critical role of detail-centric observation.

Future work will address several key areas. First, we will improve data  collection efficiency by refining teleoperation alternatives -- e.g., using AR glasses and data gloves -- to capture human active perception without robot teleoperation. Second, we will enhance viewpoint control by upgrading the gimbal mechanism and the VR-to-camera mapping to better track head motion, and by integrating additional sensors (e.g., wrist-mounted cameras) for richer observations. Finally, on policy learning, we will introduce instruction-conditioned (language-conditioned) policies to guide active perception.




\end{document}